\documentclass[11pt,a4paper]{article}
\usepackage[acceptedWithA]{tacl2018v2}
\usepackage{times,latexsym}
\usepackage{url}
\usepackage[T1]{fontenc}

\usepackage{etoolbox}
\usepackage{amssymb}
\usepackage{amsmath}
\usepackage{xcolor}%
\usepackage{comment}
\usepackage{latexsym}
\usepackage{enumitem}
\usepackage{graphicx}
\usepackage{amssymb}
\usepackage{amsmath}
\usepackage{appendix}
\usepackage{multirow}
\usepackage{booktabs}
\usepackage{subfig}
\usepackage{booktabs}
\usepackage{array}
\usepackage{algorithm}
\usepackage{algpseudocode}
\usepackage{mathtools}

\usepackage{microtype}

\newtoggle{draft}
\toggletrue{draft}
\iftoggle{draft}{
    \newcommand{\hh}[1]{\textcolor{blue}{[#1 \textsc{-HH}]}}
    \newcommand{\gn}[1]{\textcolor{red}{[#1 \textsc{-GN}]}}
    \newcommand{\pb}[1]{\textcolor{blue}{[#1 \textsc{-PB}]}}
    \newcommand{\pw}[1]{\textcolor{magenta}{[#1 \textsc{-PW}]}}
}{
    \newcommand{\hh}[1]{}
    \newcommand{\gn}[1]{}
    \newcommand{\pb}[1]{}
    \newcommand{\pw}[1]{}
}

\title{WikiAsp: A Dataset for Multi-domain Aspect-based Summarization}

\author{Hiroaki Hayashi$^\textbf{1}$ \quad Prashant Budania$^\textbf{2}$ \quad Peng Wang$^\textbf{2}$\\
        {\bf Chris Ackerson$^\textbf{2}$ \quad Raj Neervannan$^\textbf{2}$ \quad Graham Neubig$^\textbf{1}$} \\\\
  $^1$Language Technologies Institute  \quad $^2$AlphaSense\\
  Carnegie Mellon University \quad \phantom{AlphaSense}\\
  \texttt{\{hiroakih,gneubig\}@cs.cmu.edu}\\
  \texttt{\{pbudania,pwang,cackerson,rneervannan\}@alpha-sense.com} \\}

\date{}

\begin{document}
\maketitle
\begin{abstract}
    Aspect-based summarization is the task of generating focused summaries based on specific points of interest. Such summaries aid efficient analysis of text, such as quickly understanding reviews or opinions from different angles.
    However, due to large differences in the type of aspects for different domains (e.g., sentiment, product features), the development of previous models has tended to be domain-specific.
    In this paper, we propose WikiAsp,\footnote{\url{http://github.com/neulab/wikiasp}} a large-scale dataset for multi-domain aspect-based summarization that attempts to spur research in the direction of \emph{open-domain} aspect-based summarization.
    Specifically, we build the dataset using Wikipedia articles from 20 different domains, using the section titles and boundaries of each article as a proxy for aspect annotation.
    We propose several straightforward baseline models for this task and conduct experiments on the dataset.
    Results highlight key challenges that existing summarization models face in this setting, such as proper pronoun handling of quoted sources and consistent explanation of time-sensitive events.
\end{abstract}

\section{Introduction}

Aspect-based summarization is a subtask of summarization that aims to provide targeted summaries of a document from different perspectives~\cite{titov2008joint,lu2009rated,wang2016neural,yang2018aspect,angelidis2018summarizing}.
Unlike generic summarization, this gives more concise summaries that are separated according to specific points of interest, allowing readers to fulfill focused information needs more easily and quickly.
However, existing aspect-based summarization work is somewhat narrowly focused; for example a great majority of the work focuses specifically on the domain of product or restaurant reviews. %
In contrast, generic summarization models are tested on a much wider variety of genres, from newswire~\cite{nallapati2016abstractive,grusky2018newsroom}, to academic papers~\cite{kang2018dataset,kedzie2018content}, to movie scripts~\cite{gorinski2015movie}.
For each genre, the types and characteristics of aspects that will need to be touched upon in a good summary will differ greatly.

\begin{figure}[tb]
    \centering
    \includegraphics[width=\linewidth]{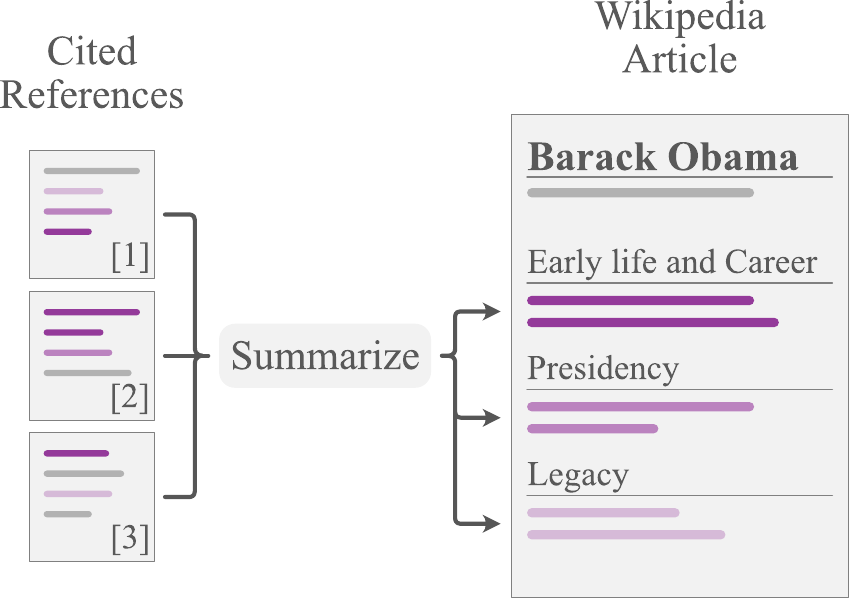}
    \caption{In WikiAsp, given reference documents cited by a target article, a summarization model must produce targeted aspect-based summaries that correspond to sections.}
    \label{fig:task}
    \vspace{-5mm}
\end{figure}

One natural source of such multi-domain articles is Wikipedia, and the section boundaries and titles in each article form natural annotations of aspects and corresponding text.
There have recently been a number of attempts to generate the \textit{lead} section of Wikipedia articles from the linked external sites in the reference section \citep{liu2018generating,fan2019using,liu2019hierarchical}, an approach that does not explicitly consider the different aspects covered by the article.
\citet{perez-beltrachini2019generating} also examine domain differences in Wikipedia text summarization.
However, existing datasets and analyses lack structure, broad domain coverage, or both.
We argue that (1) generating \emph{structured} summaries is of inherent interest, as these will allow humans consuming the information to browse specific aspects of interest more readily, and (2) the structure will \emph{vary across domains}, with different domains demonstrating very different characteristics.

In this paper, we construct a dataset for multi-domain aspect-based summarization that allows us to train models for this unique variety of summarization task, and examine the challenges posed therein.
Figure~\ref{fig:task} illustrates the overview of our task.
Specifically, we turn to \textit{section titles} of Wikipedia articles and construct sets of ``aspects'' through steps of automatic extraction, curation, and filtering.
The section texts then serve as corresponding aspect-based summaries.

We devise a baseline two-stage method consisting of aspect identification and summarization using extractive and abstractive models, and conduct experiments on the proposed dataset.
The analysis of experimental results and the generated summaries reveals the unique challenges posed by our multi-domain and multi-document setting.
For example, aspects that require summarizing contents in a particular order (\textit{e.g.}, time series events) in a multi-document setting adds extra difficulty because of the need for correctly ordering scattered (and possibly duplicate) pieces of information from different sources.
Certain domains that involve interviews or quotes of people also exhibit challenges in correctly modifying pronouns based on the relationship to the topic of interest.

\section{Generating Wikipedia as Aspect-based Summarization}
\label{sec:task}

\begin{table}
    \small
    \centering
    \begin{tabular}{@{}p{\linewidth}@{}}
        \toprule
        \textbf{Title: Barack Obama}\\\midrule
        Aspect: \textit{Early life and Career}\\\midrule
        Obama was born on August 4, 1961, at Kapiolani Medical Center for Women and Children in Honolulu, Hawaii. $\ldots$\\\midrule
        Aspect: \textit{Presidency}\\\midrule
        The inauguration of Barack Obama as the 44th President took place on January 20, 2009. In his first few days in office, Obama issued $\ldots$\\\midrule
        Aspect: \textit{Legacy}\\\midrule
        Obama's most significant legacy is generally considered to be the Patient Protection and Affordable Care Act (PPACA), $\ldots$\\
        \bottomrule
    \end{tabular}
    \caption{\label{tab:data_example}Example Wikipedia article about Barack Obama. Our goal is to generate texts given the cited references and the specified aspects.}
\vspace{-6mm}
\end{table}

Wikipedia articles exhibit a specific way of organizing information about a focused topic.
An article $S$ consists of two parts: section titles $a$, and their contents $p$.
The contents are further split into sections, where each section describes information about the main topic from different viewpoints.
Table~\ref{tab:data_example} shows an example article about the topic ``Barack Obama'', with several sections ``Early life and Career,'' ``Presidency,'' and ``Legacy''.
In practice, the contents included in each section can take many forms, from text, tables, and images, to more specialized content such as brackets of a tournament.
In this work, we focus only on sections that mainly consist of textual content (see Section~\ref{sec:data} for how we define this).

Importantly, the content in Wikipedia articles is required to be \textit{verifiable}: ``other people using the encyclopedia can check that the information comes from a reliable source''.\footnote{https://en.wikipedia.org/wiki/Wikipedia:Verifiability}
To ensure this, articles contain citations from a set of references $\mathcal{R}$ so that readers can check the validity of the content.
In other words, citations supposedly contain the majority of the information written in the articles.
\citet{liu2018generating} took advantage of this fact by proposing a summarization task using cited references as source documents for summarization.
Citations include published material (such as books) and websites, but because only web-based citations can easily and automatically be mined via crawling, we consider only web-based citations as source documents in this work and ignore the rest of non-web based citations following \citet{liu2018generating}.

The goal of our task is to learn a model $f: \mathcal{R} \rightarrow S$, which can 1) identify and gather information from cited references and 2) generate a section-by-section summary where each section contains the appropriate type of information.
Formally, let $\mathcal{R} = \{R_1, R_2,\ldots, R_{M}\}$ be a collection of $M$ cited references for an article $S = \{s_1, s_2,\ldots,s_{N}\}$ of $N$ sections.
Each section $s_i$ is essentially a tuple of a section title and one or more paragraphs: $s_i = \langle a_i, p_i \rangle$.

While there is a fair amount of variety in section titles across different articles, articles that belong to the same domain tend to share aspects that are particularly salient for that domain.
Because of this, we select a fixed-size subset of all section titles that appear in each domain as the set of aspects $\mathcal{A}$ that we will target; details on how we select this subset will be elucidated in the following section.
Hence, our task is cast as multi-document aspect-based summarization.

\section{The \textsc{WikiAsp} Dataset}
\label{sec:data}
\vspace{-2mm}
In this section, we describe our concrete steps to create our dataset.

\subsection{Data Collection}
As the base data, we build upon the data collection strategy from the WikiSum dataset~\cite{liu2018generating}, a dataset for generating lead sections of Wikipedia from referenced web pages.
Following the WikiSum data generation script,\footnote{\href{https://github.com/tensorflow/tensor2tensor/blob/master/tensor2tensor/data_generators/wikisum/README.md}{Tensor2tensor's WikiSum generator} was used.}
we first crawled cited references covered by CommonCrawl for each Wikipedia article.
We then recover all the sections\footnote{Due to the design of WikiSum dataset, the first section title of any article is automatically renamed to ``LEAD''. Therefore, we could not recover first sections of the Wikipedia articles. We suggest editing the data generation scripts for future WikiSum users if section title information is necessary.} of the target Wikipedia articles from WikiSum (which was unused in WikiSum dataset) and obtain pairs of (section title, section paragraph).
An example for this is shown in Table~\ref{tab:data_example}.

\begin{table}[tb]
    \centering
    \small
    \begin{tabular}{@{}lr|lr@{}}
    \toprule
     \multicolumn{2}{c}{Infrastructure} & \multicolumn{2}{c}{Software} \\
    \midrule
     history & 13293 & reception & 8196 \\
     route description & 5627 & gameplay & 8095  \\
     facilities & 2792 & development & 3983 \\
     services & 1955 & plot & 3697 \\
     future & 784 & history & 2465 \\
     route & 689 & features & 1799 \\
     location & 613 & story & 991  \\
     construction & 577 & release & 750 \\
     connections & 497 & overview & 570 \\
     description & 463 & legacy & 564 \\
    \bottomrule
    \end{tabular}
    \caption{\label{tab:aspstats_4}Frequency of filtered aspects that are \textit{textual} in 2 domains. Due to space constraint, the statistics for the rest of domains will be available in the Appendix~\ref{app:aspect_stats}.}
\vspace{-5mm}
\end{table}

\subsection{Domain Separation}
Articles in different domains focus on different salient topics, as observed by~\citet{perez-beltrachini2019generating}. %
For example, the ``discography'' section is common for articles about singers, but is not appropriate for articles about infrastructure.
To characterize such structural differences, we separate the set of articles obtained in the previous step into sets in particular domains.
Specifically, we follow \citet{perez-beltrachini2019generating} in assigning one category for each article using DBPedia~\cite{auer2007dbpedia}.
DBPedia stores structured information for each Wikipedia article, including the domain labels and info boxes.
Additionally, it defines a topical hierarchy of the domains (ontology classes).
We first map between articles and the domain labels from the corresponding DBPedia dump.
Obtained domain labels, however, have mixed granularity (e.g., Person and its sub-class Dancer) which causes imbalance in the number of examples in each domain, as well as domain overlap between high-level and low-level domains in the domain hierarchy.
We mitigate this by recursively merging domains at leaf-level into coarser ones according to the aforementioned topical hierarchy from the ontology classes.\footnote{http://mappings.dbpedia.org/server/ontology/classes/}
We repeat the merging procedure until a branch in the hierarchy includes more than 15,000 articles, and picked 20 domains at the leaf of the merged hierarchy.\footnote{Many articles are labeled directly as Person, in which case the domain is high-level at the hierarchy. We do not select this domain because lower-level domains such as Artist or SoccerPlayer already have enough number articles.}

\begin{table*}[tb]
    \centering
    \small
    \begin{tabular}{@{}llrrrrrr@{}}
    \toprule
    Dataset                                & Domain & \#Dom. & \#Train & Doc. Length & Sum. Length & \#Asp. & \#Asp./Ex. \\\midrule
    OpoSum & Product Review & 6 & 359,048        & 138  & 49 & 9    & 2.00 \\
    Amazon & Product Review & 7 & 240,000        & 82   & -  & -    & - \\
    RottenTomatoes& Movie Review   & 1 & 2,458          & 2369 & 24 & $^\ast$2 & $^\ast$1.00 \\
    MA-News & News           & 1 & 284,701        & 1350 & 54 & 6  & 2.98 \\\midrule
    \textbf{WikiAsp}                                & Encyclopedia   & 20& 320,272        & 13,672 & 213 & 10 & 1.77 \\\bottomrule
    \end{tabular}
    \caption{\label{tab:data_compare}Training set statistics comparisons against previous aspect-based summarization datasets. For multi-domain datasets, the sum of all the examples are reported. \#Asp./Ex. represents the average number of aspects that a model has to summarize on each example. ($^\ast$ Review saliency is treated as aspects. \#Asp. represents the number of aspects per domain if the number of domains is more than one. Compared datasets are the work of \citet{angelidis2018summarizing,yang2018aspect,wang2016neural,frermann2019inducing}, respectively.}%
    \vspace{-5mm}
\end{table*}

\subsection{Aspect Selection}

Next, we perform aspect selection on each set of articles in the domains extracted during the previous step.
As previously noted, articles in the same domain tend to share similar set of section titles.
Motivated by this observation, we construct the set of aspects from the \textbf{most frequent section titles}.

From the frequency distribution of section titles in a domain, we manually filter ones that are not \textit{textual}, that is, more than half portion of section consists of text.
For each section title, we take 20 randomly sampled sections and include it in the set of aspects only if 80\% of samples consist of \textit{textual} paragraphs.
Following the steps above, we construct the 10 most frequent aspects for each domain.
However, the choice of words in section titles vary depending on the editors within the same domain, which leads to missing relevant aspects that are moderately frequent but not present in Top-10.
For example, one of the common section titles in WrittenWork domain are ``summary'' and ``plot summary,'' which should be merged together to form a single aspect.
We handle these cases by inspecting the frequent distribution further down and manually identifying semantically equivalent titles to merge.

The resulting dataset consists of instances in 20 domains where each domain has 10 pre-defined aspect classes.
We show statistics comparisons of the dataset to existing aspect-based summarization datasets in Table~\ref{tab:data_compare} and examples of obtained aspects for two domains in Table~\ref{tab:aspstats_4}.

Appendix~\ref{app:domain_stats} and \ref{app:aspect_stats} summarizes the data size for each domain and the obtained aspects for the rest of 18 domains respectively.

\begin{figure*}[t]
    \centering
    \includegraphics[width=0.8\linewidth]{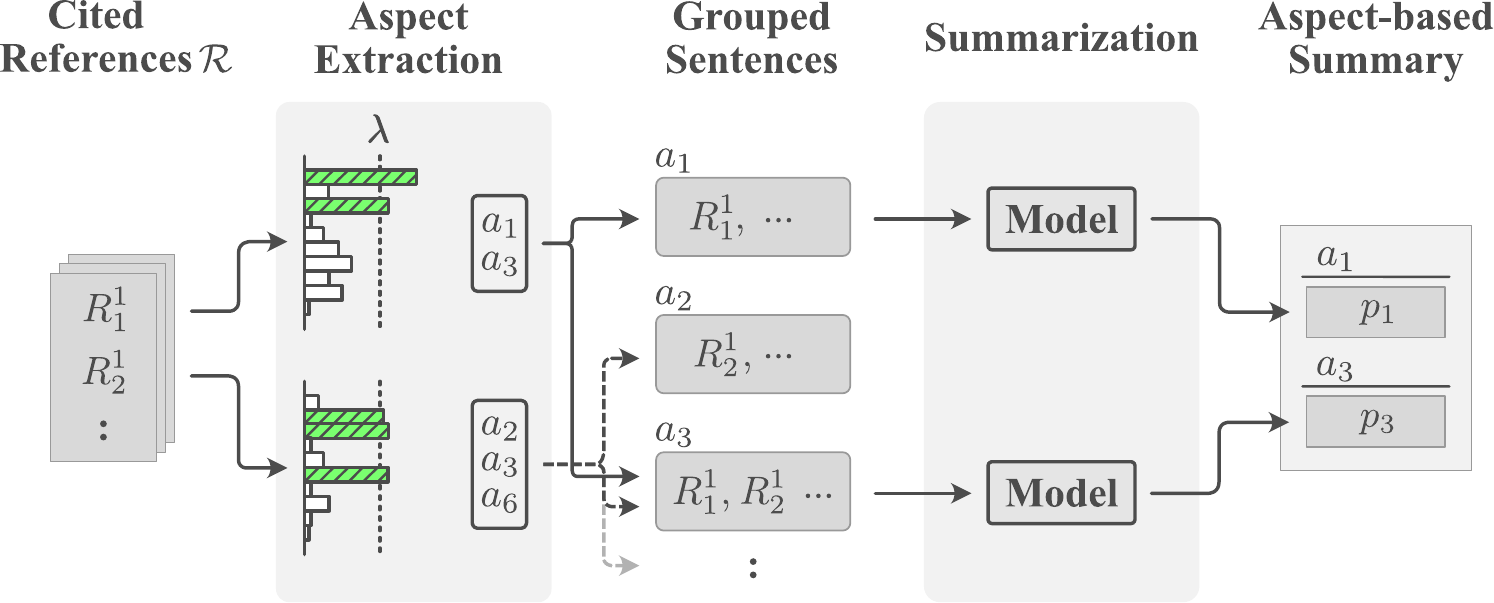}
    \caption{Two-stage model diagram. The aspect classifier assigns aspect labels for each reference sentence $R^i_j$ from references $\mathcal{R}$ with a threshold $\lambda$. Sentences are then grouped according to the assigned labels, which are fed to the summarization model. Groups about irrelevant aspects (\textit{i.e.}, $a_2$) is ignored. Finally, the summarization model outputs summaries for each relevant aspect.}
    \label{fig:model}
    \vspace{-5mm}
\end{figure*}

\section{Baseline Models}
\label{sec:model}

Next, in this section we describe two baseline models for solving this task.
Both of these models decompose the overall process into two stages: aspect discovery and aspect-based summarization of classified sentences.
Both baseline models share the same methodology for aspect discovery, but differ in terms of summarization models.
The model overview is shown in Figure~\ref{fig:model}.

\subsection{Aspect Discovery}
The first stage consists of labeling sentences in \textit{cited reference texts} according to aspects.
Having training data that contains sentences in the reference documents labeled with target aspects would be the ideal case, but these do not exist \emph{a priori}.
Therefore, we instead create training data by assigning each sentence in the target articles with aspect labels corresponding to the aspect to which the sentence belongs. %
For example, the article about Barack Obama in Table~\ref{tab:data_example} yields training instances consisting of sentences labeled with \textit{Early life and career}, \textit{Presidency} and \textit{Legacy} depending on which paragraph a sentence comes from.
This data makes it possible to train a classifier that learns to predict aspects from the texts at sentence-level.
At test time, cited reference sentences are fed into the learned classifier and are labeled with their most likely aspects.

However, the discrepancy of inputs at train/test time is problematic because the model is not exposed to any \textit{noisy} sentences that do not belong to any of the relevant aspects at training time, while cited reference texts do contain such sentences.
For example, an article in the \textit{Company} domain may have a citation to the company website itself, which contains commercial messages that may not be appropriate in encyclopedic text such as Wikipedia. %
We manage such cases by introducing an auxiliary label \textit{Other} at training time and let the model learn to identify noisy sentences as well.
To do so, sentences labeled with \textit{Other} are randomly sampled from texts in different domains and added to training data.
We fine-tune the pre-trained ROBERTa~\cite{liu2019roberta} model on this classification dataset for each domain.
Logits obtained from the model are then passed through the sigmoid function to obtain probabilities of each aspect for a given sentence.
Finally, we assign labels to a sentence by taking the aspects $a_i$ whose probabilities are greater than the threshold $\lambda$: $P(a_i) > \lambda$.
The lower we set the threshold, the more but potentially noisy sentences we include as the input to the summarization model. %
We tune $\lambda$ independently for each domain based on the performance on validation sets and set $0.5$ for \textit{Group}, $0.8$ for \textit{Album}, \textit{Animal}, \textit{Building}, \textit{Film}, and $0.9$ for the remaining domains as the threshold values.

\subsection{Summarization}
Sentences that are labeled with the same aspect are then grouped in order of occurrence in cited references to form a chunked paragraph that discusses the same aspect.
This forms aspect-based clusters of relevant sentences, which become the input to a summarization model.
On the contrary, aspects that are never labeled (due to low probabilities) are deemed irrelevant and thus will not be summarized.
We consider both an extractive and an abstractive summarization model in our baseline implementation.
For the extractive model, we use TextRank~\cite{mihalcea2004textrank,BarriosLAW16}, a graph-based ranking model for extracting important sentences.
For the abstractive model, we use PreSumm~\cite{liu2019text}, a Transformer-based summarizer with fine-tuned BERT as the source encoder.
For each domain, PreSumm is fine-tuned and trained on the pairs of (grouped sentences, target aspect paragraph) to learn to produce summaries given the aspect-relevant sentences.

\section{Evaluation}
\vspace{-2mm}
We evaluate models along two axes: aspect discovery and summarization.
We note that the primary task in this dataset is aspect-based summarization, thus aspect discovery evaluation discussed below is only for diagnostic purposes.
Since the aspect sets differ in different domains, evaluation is performed separately for each domain.

\paragraph{Aspect Discovery}
Models have to correctly predict the right set of aspects about which they generate summaries.
The aspect discovery criterion aims to evaluate the similarity between the set of aspects about which a model decides to generate summaries and the set of aspects that appear in the target article.%
\footnote{Note that there are two potential reasons an aspect does not appear in the target article: (1) it may not be appropriate for that particular entity (e.g. the ``controversy'' aspect in the ``company'' domain should not exist if that company has legitimately never had a controversy), or (2) the article may not be complete. For this evaluation, we make the simplifying assumption that all articles are complete and thus missing aspects are an indication of failure to recall information, but relaxing this assumption in some way may result in more accurate evaluation.}
For comparing these two sets, we use precision, recall and F1 scores.

\vspace{-2mm}
\paragraph{Aspect-based Summarization}
Gold standard summaries only exist for each of the aspects that appear in an article.
Therefore in this evaluation, we focus on evaluating the model's ability to summarize inputs particularly on these aspects.
Specifically, generated summaries are paired to corresponding reference summaries with the same aspects and are evaluated using ROUGE~\cite{lin2004rouge}.
Since ROUGE is a recall-based measure, the number of tokens in the model outputs directly affect the performance.
Controlling the length is particularly important for our dataset because average summary length for each aspect in different domains varies (\textit{e.g.}, ``description'' and ``location'' from HistoricPlace domain has 396 and 90 average tokens, respectively).
We take this into account by explicitly setting the maximum number of words for extractive and abstractive summaries to be the average number of words in the target summaries in the training set for each aspect and for each domain.

\section{Experiments}
\vspace{-2mm}
We provide two baseline models for the task and evaluate on the proposed dataset.

\subsection{Implementation Details}
\label{exp:implementation}
For aspect classification, we used \texttt{roberta-base}\footnote{We used Huggingface's implementation~\cite{Wolf2019HuggingFacesTS} for obtaining and fine-tuning the weights.} model and fine-tuned for 5 epochs on the created surrogate dataset above for each domain, with the learning rate $2 \times 10^{-5}$.
For the extractive summarization, we specify the summary length for TextRank according to the mean length of target summaries for each aspect in each domain.
We re-train the PreSumm summarizer on our dataset for each domain: the encoder is initialized with the weights of pre-trained BERT~\cite{devlin2019bert} and the decoder is trained from scratch.
The total number of training steps is 300,000.
For some domains, we further tuned the decoder dropout rate to $0.3$ to stabilize training.
At inference time, we specify maximum summary lengths for each \textit{aspect} for each domain using the average summary lengths from computed from the training set.

\subsection{Results}
In this section, we discuss the experimental results on each stage.

\subsubsection{Aspect Discovery}
We show the aspect discovery results in Table~\ref{tab:asp_discovery}.
We see a general trend of high recall predictions made by the model.
While varying thresholds could balance precision and recall, the results exhibited high recall after hyperparameter search.
This suggests that the learned classifier is poorly calibrated.
Class imbalance also plays a role here; predicting the major classes give high recall due to skew aspect frequency distributions.
Among others, the classifier performed best with the Town domain by achieving the highest precision and the F1 score.

\begin{table}[tb]
    \small
    \centering
    \begin{tabular}{@{}lrrr@{}}
    \toprule
    Domain & Prec & Rec & F-1 \\\midrule
    Album & 19.64 & 86.43 & 30.64 \\
    Animal & 34.69 & 84.08 & 45.52 \\
    Artist & 26.32 & 75.24 & 36.72 \\
    Building & 31.46 & 91.25 & 42.92 \\
    Company & 28.97 & 91.50 & 41.06 \\
    EducationalInstitution & 25.64 & 93.82 & 37.66 \\
    Event & 28.99 & 96.44 & 42.36 \\
    Film & 32.84 & 91.46 & 45.17 \\
    Group & 17.46 & 95.56 & 28.18 \\
    HistoricPlace & 33.38 & 90.22 & 42.98 \\
    Infrastructure & 28.38 & 94.00 & 41.00 \\
    MeanOfTransportation & 23.24 & 83.13 & 33.88 \\
    OfficeHolder & 21.22 & 73.25 & 30.62 \\
    Plant & 31.25 & 83.17 & 42.10 \\
    Single & 25.36 & 88.33 & 37.16 \\
    SoccerPlayer & 28.54 & 67.18 & 37.16 \\
    Software & 31.52 & 94.65 & 45.10 \\
    TelevisionShow & 20.44 & 81.76 & 31.28 \\
    Town & 42.61 & 71.85 & 50.12 \\
    WrittenWork & 21.50 & 94.29 & 33.71 \\
    \bottomrule
    \end{tabular}
    \caption{Aspect discovery results on the test set.\label{tab:asp_discovery}}
    \vspace{-4mm}
\end{table}

\subsubsection{Summarization}
The automatic evaluation results are shown in Table~\ref{tab:summ}.
Neither baseline unanimously outperformed the other on all domains, but we observe that PreSumm (abstractive) performs better than TextRank (extractive) on average.%
The low R-2 and R-L scores by both models despite the oracle being relatively higher suggest that important phrases to be summarized do not appear rarely.\footnote{Note that TextRank connects nodes according to content overlap, thus isolated sentences are not selected.}

To understand the upper-bound of model performance for the task, we also show summarization results of the extractive oracle model in Table~\ref{tab:summ}.
Sentences were chosen directly from cited reference texts to maximize the ROUGE score against summaries, thus bypassing the aspect classification stage.
The oracle performance shows that a summarization model can indeed perform competitively on the dataset if the model is given with the full input information.
The contrasting results between the oracle and two stage models suggests the importance of accurate content selection before performing summarization.

\begin{table*}[tb]
    \centering
    \small
    \begin{tabular}{@{}lrrr|rrr|rrr@{}}
    \toprule
    & \multicolumn{3}{c}{TextRank} & \multicolumn{3}{c}{PreSumm} &\multicolumn{3}{c}{Extractive Oracle} \\\cmidrule{2-4}\cmidrule{5-7}\cmidrule{8-10}
    & R-1 & R-2 & R-L & R-1 & R-2 & R-L & R-1 & R-2 & R-L\\\midrule
    Album & 19.56 & 2.81 & 17.26 & 22.76 & 6.31 & 20.27 & 37.72 & 12.58 & 33.19 \\
    Animal & 18.00 & 3.16 & 16.05 & 27.11 & 8.08 & 25.01 & 34.82 & 10.52 & 31.01 \\
    Artist & 17.22 & 2.49 & 15.58 & 21.79 & 3.76 & 20.00 & 41.49 & 15.04 & 37.64 \\
    Building & 23.91 & 4.96 & 21.85 & 24.99 & 5.97 & 23.24 & 41.95 & 14.31 & 38.28 \\
    Company & 22.92 & 3.70 & 20.65 & 22.28 & 4.08 & 20.50 & 40.20 & 12.30 & 36.16 \\
    EducationalInstitution & 21.47 & 4.29 & 19.24 & 24.17 & 6.70 & 21.96 & 39.11 & 14.04 & 35.18 \\
    Event & 26.64 & 5.67 & 24.08 & 28.31 & 7.69 & 26.20 & 46.17 & 16.90 & 41.87 \\
    Film & 21.25 & 3.81 & 19.14 & 20.58 & 5.34 & 18.86 & 40.24 & 13.78 & 36.14 \\
    Group & 22.30 & 3.62 & 20.20 & 25.51 & 4.97 & 23.51 & 41.36 & 13.23 & 37.56 \\
    HistoricPlace & 18.96 & 3.71 & 17.51 & 27.40 & 8.08 & 25.69 & 37.78 & 10.83 & 34.65 \\
    Infrastructure & 20.40 & 3.27 & 18.39 & 27.86 & 9.24 & 25.80 & 36.04 & 10.00 & 32.25 \\
    MeanOfTransportation & 21.20 & 3.93 & 19.31 & 24.52 & 7.04 & 22.72 & 41.13 & 13.70 & 37.45 \\
    OfficeHolder & 18.45 & 3.15 & 16.77 & 19.63 & 5.24 & 18.12 & 39.60 & 14.70 & 36.04 \\
    Plant & 18.73 & 3.02 & 16.84 & 25.29 & 6.30 & 23.20 & 34.93 & 9.66 & 31.31 \\
    Single & 17.96 & 2.67 & 15.86 & 22.06 & 6.78 & 19.98 & 36.51 & 11.57 & 31.88 \\
    SoccerPlayer & 14.79 & 2.36 & 12.89 & 12.89 & 1.86 & 12.05 & 31.06 & 8.00 & 27.08 \\
    Software & 24.54 & 4.56 & 22.05 & 20.51 & 5.15 & 18.82 & 42.79 & 13.96 & 38.30 \\
    TelevisionShow & 19.77 & 3.21 & 17.68 & 19.20 & 3.53 & 17.42 & 40.35 & 13.47 & 35.67 \\
    Town & 17.89 & 3.56 & 16.50 & 19.76 & 4.39 & 16.87 & 33.21 & 10.31 & 30.70 \\
    WrittenWork & 23.39 & 3.89 & 21.14 & 22.19 & 4.33 & 20.15 & 42.66 & 13.93 & 38.16 \\\midrule
    AVG & 20.47 & 3.59 & 18.45 & 22.94 & 5.74 & 21.02 & 38.95 & 12.64 & 35.03 \\
    \bottomrule
    \end{tabular}
    \caption{Aspect-based summarization results on the test set. The last row shows the average performance.\label{tab:summ}}
\end{table*}

\section{Analysis}
We discuss the model outputs and analysis below.

\begin{table}[tb]
    \small
    \centering
    \begin{tabular}{@{}llrr@{}}
    \toprule
    \multirow{2}{*}{Dom.} & \multirow{2}{*}{Aspect} & PreSumm & TextRank \\\cmidrule{3-4}
         &         & $\downarrow$ R-1 & R-1 \\\midrule

    Tow. & government & \textbf{55.10} & 21.20 \\
    Eve. & format & 44.94 & 24.73 \\
    Inf. & facilities & 42.46 & 14.75 \\
    Bui. & exterior & 41.81 & 25.60 \\
    Mea. & background & 39.00 & 23.72 \\
    His. & heritage listing & 36.58 & 10.25 \\
    Ani. & habitat & 32.91 & 12.95 \\
    Pla. & taxonomy and nm. & 32.70 & 9.39 \\
    Edu. & rankings & 31.80 & 26.92 \\
    Alb. & commercial perf. & 31.71 & 15.51 \\\midrule
    Dom. & Aspect & R-1 &  $\downarrow$ R-1 \\\midrule
    Eve. & battle & 28.00 & \textbf{32.00} \\
    Eve. & report & 24.77 & 30.11 \\
    Sof. & gameplay & 24.17 & 28.53 \\
    Eve. & background & 30.01 & 27.42 \\
    Eve. & aftermath & 27.54 & 27.27 \\
    Bui. & history & 25.32 & 27.13 \\
    Sof. & plot & 20.50 & 27.00 \\
    Edu. & rankings & 31.80 & 26.92 \\
    Wri. & plot summary & 22.08 & 26.85 \\
    Fil. & plot & 19.43 & 26.66 \\\bottomrule

    \end{tabular}
    \caption{\label{tab:sorted_asp_rouge}List of aspects sorted in descending order of ROUGE-1 score according to PreSumm (top half) and TextRank (bottom half). ``performance'' and ``naming'' are abbreviated to ``perf.'' and ``nm.'', respectively. Domain names shortened to the first three letters.}
\end{table}

\subsection{Aspect-by-aspect Evaluation}
Not all the aspects are equally hard to summarize; some might require summarization of a broad range of information, while others require only specific concepts to be summarized.
We further investigate this by looking into summarization performance for both models on per-aspect basis.
Table~\ref{tab:sorted_asp_rouge} shows the best-performing aspects sorted in descending order by ROUGE-1 scores for two summarization models on the validation set.
Through manual investigation of the generated samples for each aspect, we observed that the aspects where the \textit{abstractive} model performed well  tend to have common templates and similar choice of vocabulary, more so than other aspects.
For example, 58\% (out of 183 samples) of the target summaries for \textit{government} in Town shared the identical summaries despite the fact that articles discuss different townships.
Similar but less prevalent patterns were observed in other aspects as well.

Aspects where the \textit{extractive} summarization model performed better contain much larger numbers of tokens in the summaries than average.
Specifically, the average summary length for 10 aspects where TextRank performed the best was 303, while that for 10 aspects where PreSumm performed the best was 166.
Naturally, abstractive models have issues with maintaining coherence over long decoding results, but the extractive model has few issues gathering relevant sentences at the cost of incoherent transitions from sentence to sentence.
As for the content, extractive summaries exhibited the advantage of being able to correctly include mentions related to numbers and dates.

\begin{table*}[tb]
    \small
    \centering
    \begin{tabular}{@{}p{\linewidth}@{}}
        \toprule
        \textbf{Domain / Title: Software / Cyberpunk 2077 }\\\midrule
        Aspect: \textit{Gameplay}\\\midrule
        \textbf{Gold}: cyberpunk 2077 is a role - playing video game played from either a first - person or third - person perspective . it is set in an open world metropolis called night city . the game will feature non - english speaking characters . players who do not speak the languages can buy translator implants to better comprehend them ; $\ldots$\\
        \textbf{Ext.}: cyberpunk 2077 takes place in , you guessed it , the year 2077 . for just a few hours , you can be rich , successful , and popular with your preferred gender . cyberpunk 2077 will be a aaa rpg , but whether it will come to resemble the witcher in any way remains unclear . how braindances will be used by or on the protagonist is $\ldots$ \\
        \textbf{Abs.}: the game is an action role - playing game in which the player assumes the role of an unnamed protagonist , runabout who lives in a large metropolis known as the `` black city `` . the game ' s narrative is told through procedurally generated procedurally generated . during the course of the game , the player is briefed into $\ldots$\\
        \toprule
        \textbf{Domain / Title: Event / Battle of the Saintes }\\\midrule
        Aspect: \textit{Aftermath}\\\midrule
        \textbf{Gold}: the british lost 243 killed and 816 wounded , and two captains out of 36 were killed . the french loss in killed and wounded has never been stated , but six captains out of 30 were killed . it is estimated that the french may have lost as much as 3 , 000 , and more than 5 , 000 french soldiers and sailors were captured , $\ldots$\\
        \textbf{Ext.}: battle of the saintes , ( april 9 – 12 , 1782 ) , in the american revolution , major naval victory for britain in the west indies that restored british naval mastery in the area and ended the french threat to nearby british possessions . the valiant was subsequently present at the relief of gibraltar on 12 april 1781 $\ldots$\\
        \textbf{Abs.}: the immediate aftermath of the battle of the niles meant the british fleet could not be able to deliver a great strategic victory in the kingdom of france , and had also declared war on great britain with the royal navy to france . the outcome was the subject of a strategic victory for the british .  in the meantime , the british navy had captured the french navy $\ldots$ \\ %
        \midrule
        Aspect: \textit{Battle}\\\midrule
        \textbf{Gold}: on 9 april 1782 , the copper - hulled british fleet soon caught up with the french , who were surprised by their speed . de grasse ordered the french convoy to head into guadeloupe for repair , forcing him to escort two fifty - gun ships ( fier and experiment ) and placing his fleet in line of battle in order to cover the retreat . $\ldots$ \\
        \textbf{Ext.}: " after some initial maneuvers and minor clashes , a full - scale battle was joined on april 12 , by which time the british had thirty - six ships of the line in action against thirty french ones . he turned his ships ninety degrees and sailed through the broken french line of battle , splitting the french line into four segments . in doing this , the guns on each side of the british ships were brought to bear on the french with little risk of return fire . $\ldots$\\
        \textbf{Abs.}: the battle started as the shock . the battle progressed , when the british forces reached the north - eastern flank of the battle of weidman ( kingdom ) in a battle . he had begun to return to the field , and moved his forces toward the mouth of the river . in the battle , the first contingent of the french navy ships got off from a small contingent of british soldiers as well as the third - rate , under the command of general sir henry sturgis . $\ldots$\\
        \bottomrule
    \end{tabular}
    \caption{\label{tab:summaries}Generated summaries from multiple domains. Ext. and Abs. represent summaries from TextRank and PreSumm.}
\end{table*}

\subsection{Quality of Generated Summaries}

We then examined the generated summaries from the two models and compared them qualitatively.
Samples are shown\footnote{Samples from other domains are in Appendix~\ref{app:samples}.} in Table~\ref{tab:summaries} from some of the domains listed in Table~\ref{tab:aspstats_4}.

Manual inspection of the generated summaries revealed pros and cons of the two models:

\begin{itemize}
    \item \textbf{Both models are successful at discussing on-topic content.} For all the summaries inspected, both models were able to generate on-topic content in spite of the source documents potentially being noisy.
    \item \textbf{Abstractive summaries underperform at generating exact entity mentions.} Almost all the samples require generation of entities because the task targets at generating encyclopedic texts. Except for the title (topic) entity, abstractive models either generated no entities or wrong ones.
\end{itemize}

\subsection{Aspect Classification Accuracy}
\begin{figure}[tb]
    \centering
    \includegraphics[width=0.8\linewidth]{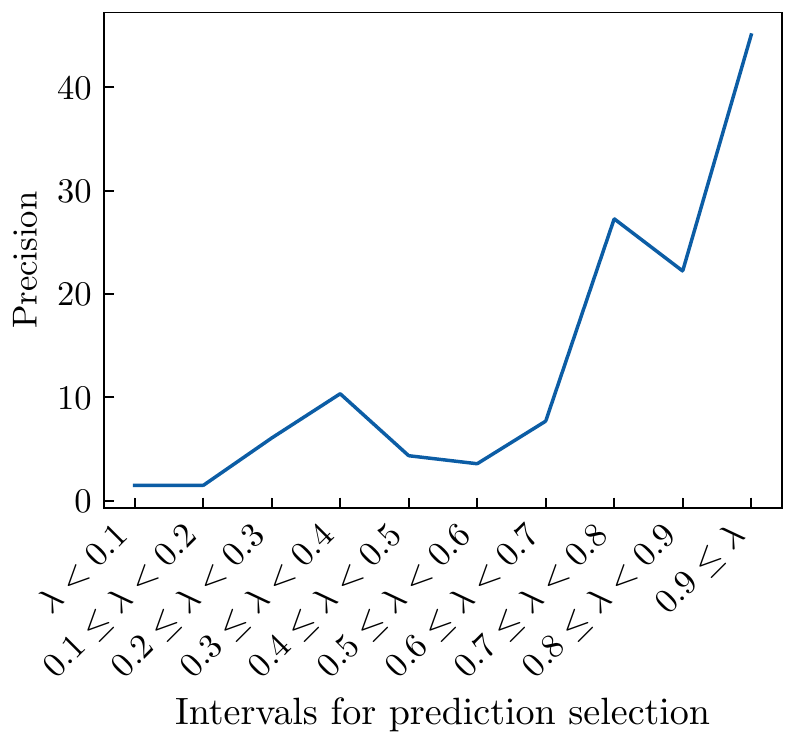}
    \caption{Precision differences in varying threshold ranges.}
    \label{fig:precision}
    \vspace{-4mm}
\end{figure}

We observed a general trend of low precision for aspect discovery.
We hypothesize that this is due to limited target aspects for each article; correctly extracted aspects affect negatively to precision if they do not exist in the target article.
To quantify this, 10 random articles are selected from the validation set in Software domain. For each article, we extract 10 sentences labeled with the highest confidence for each of the 10 aspects, resulting in 1,000 sentences in total.
Each sentence is annotated with binary labels indicating whether it is correctly associated with the aspect or not.\footnote{Sometimes, the entity in discussion by the sentence is not clear. In this case, we annotate it correct if the sentence could correspond to the target aspect of any entity.}
With the threshold $\lambda$ set to 0.9, we achieved the precision of 45.1, which shows that the aspect discovery has the ability to extract aspects, but not as good at extracting \textit{relevant} aspects for the article.
We observed that the model predictions tend to be polarized to extreme values (\textit{i.e.}, near 0 or 1).
We also show the relationship between $\lambda$ ranges and the precision in Figure~\ref{fig:precision}, which indicates that the classifier is not well-calibrated.

\subsection{Domain-specific Challenges}
One of the benefits of having many domains for the same task is to be able to characterize the differences and challenges that are unique to certain domains.
We analyzed the generated summaries from both of the summarization models and identified some of them below.

\subsubsection{Pronoun Resolution for Opinion-based Inputs}
This is particularly important in domains and aspects with subjective reviews such as music(Album, Artist, Group, and Single) or Software.
Source documents in these domains often include quotes by artists or critics, which are often written from different person perspective.
These are usually converted by the Wikipedia editors into more encyclopedic text, citing the source of the information and writing in the third person.
By design, extractive summaries have issues with this problem because of the lack of ability to transform the input sentences in any way.
For example, the first extractive summary in Table~\ref{tab:summaries} describes a game in a subjective way.
We verified this by randomly selecting 20 summaries for \textit{gameplay} aspect in Software domain.
We inspected pronouns in extractive summaries and mark ones with first- or second-person pronouns if the gold summaries do not contain them.
We found 65\% of the samples contained those undesirable pronouns that do not align with the format of gold summaries.

\subsubsection{Chronological Explanation}
This variety of content is often found in certain aspects such as \textit{history} and \textit{event}, which tend to appear across multiple domains but are most prevalent in Event, HistoricPlace, and non-human entities like Company and Building.
It is essential in these aspects to describe key information in the right chronological order for better readability.
This would not be a hard task for single document summarization, as the model could perform reasonably by following the order of the original document.
However, since our input is of multi-document form, maintaining chronological order when aggregating information across multiple domains becomes non-trivial.
Indeed, neither of the models were successful at being truthful to the order even when there are enough clues in the original references.
For example, multiple sentences start with ``In [year], $\ldots$'', but the generated summary jumps around in time.
We randomly picked 20 samples of extractive summaries with \textit{history} aspect from Company domain and found that 25\% of the samples have inconsistent timeline explanations.

\section{Related Work}

\subsection*{Aspect-based Summarization}
Aspect-based summarization has been widely investigated primarily on product or restaurant reviews~\cite{titov2008joint,lu2009rated,yang2018aspect,wang2016neural}.
\citet{angelidis2018summarizing} proposed a weakly supervised method for aspect-based opinion summarization that discovers aspects with a topic model and does not require gold aspect annotation.
TAC 2010 held a shared task of guided-based summarization on newswire domain, which resembles aspect-based summarization in terms of topic guidance.
Recently, the task has been extend to news-domain by generating artificial datasets for aspect-based summarization to address the lack of large-scale data with aspect annotation~\cite{frermann2019inducing,krishna2018generating}.
Our work also builds an aspect-based summarization dataset automatically and is most similar to~\citet{krishna2018generating}, but utilizes naturally available online encyclopedia entries and their sections in multiple domains.

\subsection*{Wikipedia as a Summarization Dataset}
Wikipedia has been studied as a target resource for generation.
An early attempt on generating \textit{full} Wikipedia articles relied on web search results for target entities as inputs~\cite{sauper2009automatically}, which simulates an authoring process of humans searching information over the Internet.
\citet{liu2018generating} formulate a sub-task of generating \textit{lead} sections as summarization of reference web pages to target articles.
The resulting WikiSum dataset is accompanied by rich metadata about articles and inspired different uses of the dataset~\cite{perez-beltrachini2019generating}.
Our work also builds upon the WikiSum dataset, and aims to evaluate aspect-based summarization models using different sections from Wikipedia articles.
Compared to~\citet{sauper2009automatically}, our dataset is an order of magnitude larger, both in the amount of articles and in the number of domains  covered.

\subsection*{Multi-Document Summarization}
Extractive methods have shown effective for multi-document summarization in previous work \cite{nenkova2006compositional,cao2015ranking,yasunaga2017graphbased}, but abstractive methods have increasingly adopted for the task~\cite{lebanoff2018adapting,fabbri2019multinewsa}.
Our task is based on the idea of \cite{liu2018generating} which treats references as source documents for the multi-document summarization task, and we experimented with both types of summarization models in our experiments.

\section{Conclusion and Future Work}%
\label{sec:conclusion}

In this paper, we propose a large-scale, multi-domain multi-aspect summarization dataset derived from Wikipedia.
Through experiments, we perform an extensive analysis of performance across different genres and aspect types.
Our analysis has demonstrated that there are both general challenges regarding summarization into various aspects, as well as specific challenges in particular genres such as time-consistent mentions and proper pronoun conversion depending on the writer of the original content.

Because of this, the proposed dataset also provides a testbed for several potential directions for future work.
For example, better aspect discovery
models may take into account the coherence of the discourse in the original documents when extracting aspects.
Better summarization models may take into account the provenance of the information, appropriately determining when the information is written by a first or third party.
WikiAsp also invites a focus on domains of interest to investigate various problems of text summarization, such as correct pronoun handling and description of chronological timeline.

\section{Acknowledgment}
We would like to thank anonymous reviewers for insightful comments. HH and GN were supported by a grant from AlphaSense.

\bibliography{acl2020}

\begin{thebibliography}{30}
\expandafter\ifx\csname natexlab\endcsname\relax\def\natexlab#1{#1}\fi

\bibitem[{Angelidis and Lapata(2018)}]{angelidis2018summarizing}
Stefanos Angelidis and Mirella Lapata. 2018.
\newblock \href {https://doi.org/10.18653/v1/D18-1403} {Summarizing
  {{Opinions}}: {{Aspect Extraction Meets Sentiment Prediction}} and {{They Are
  Both Weakly Supervised}}}.
\newblock In \emph{Proceedings of the 2018 {{Conference}} on {{Empirical
  Methods}} in {{Natural Language Processing}}}, pages 3675--3686, {Brussels,
  Belgium}. {Association for Computational Linguistics}.

\bibitem[{Auer et~al.(2007)Auer, Bizer, Kobilarov, Lehmann, Cyganiak, and
  Ives}]{auer2007dbpedia}
S{\"o}ren Auer, Christian Bizer, Georgi Kobilarov, Jens Lehmann, Richard
  Cyganiak, and Zachary Ives. 2007.
\newblock \href
  {https://link.springer.com/chapter/10.1007/978-3-540-76298-0_52} {Dbpedia: A
  nucleus for a web of open data}.
\newblock In \emph{The semantic web}, pages 722--735. Springer.

\bibitem[{Barrios et~al.(2016)Barrios, L{\'{o}}pez, Argerich, and
  Wachenchauzer}]{BarriosLAW16}
Federico Barrios, Federico L{\'{o}}pez, Luis Argerich, and Rosa Wachenchauzer.
  2016.
\newblock \href {http://arxiv.org/abs/1602.03606} {Variations of the similarity
  function of textrank for automated summarization}.
\newblock \emph{CoRR}, abs/1602.03606.

\bibitem[{Cao et~al.(2015)Cao, Wei, Dong, Li, and Zhou}]{cao2015ranking}
Ziqiang Cao, Furu Wei, Li~Dong, Sujian Li, and Ming Zhou. 2015.
\newblock \href {https://dl.acm.org/doi/10.5555/2886521.2886620} {Ranking with
  recursive neural networks and its application to multi-document
  summarization}.
\newblock In \emph{Twenty-ninth AAAI conference on artificial intelligence}.

\bibitem[{Devlin et~al.(2019)Devlin, Chang, Lee, and
  Toutanova}]{devlin2019bert}
Jacob Devlin, Ming-Wei Chang, Kenton Lee, and Kristina Toutanova. 2019.
\newblock \href {https://doi.org/10.18653/v1/N19-1423} {{BERT}: Pre-training of
  deep bidirectional transformers for language understanding}.
\newblock In \emph{Proceedings of the 2019 Conference of the North {A}merican
  Chapter of the Association for Computational Linguistics: Human Language
  Technologies, Volume 1 (Long and Short Papers)}, pages 4171--4186,
  Minneapolis, Minnesota. Association for Computational Linguistics.

\bibitem[{Fabbri et~al.(2019)Fabbri, Li, She, Li, and
  Radev}]{fabbri2019multinewsa}
Alexander Fabbri, Irene Li, Tianwei She, Suyi Li, and Dragomir Radev. 2019.
\newblock \href {https://doi.org/10.18653/v1/P19-1102} {Multi-{{News}}: {{A
  Large}}-{{Scale Multi}}-{{Document Summarization Dataset}} and {{Abstractive
  Hierarchical Model}}}.
\newblock In \emph{Proceedings of the 57th {{Annual Meeting}} of the
  {{Association}} for {{Computational Linguistics}}}, pages 1074--1084,
  {Florence, Italy}. {Association for Computational Linguistics}.

\bibitem[{Fan et~al.(2019)Fan, Gardent, Braud, and Bordes}]{fan2019using}
Angela Fan, Claire Gardent, Chlo{\'e} Braud, and Antoine Bordes. 2019.
\newblock \href {https://doi.org/10.18653/v1/D19-1428} {Using local knowledge
  graph construction to scale {S}eq2{S}eq models to multi-document inputs}.
\newblock In \emph{Proceedings of the 2019 Conference on Empirical Methods in
  Natural Language Processing and the 9th International Joint Conference on
  Natural Language Processing (EMNLP-IJCNLP)}, pages 4186--4196, Hong Kong,
  China. Association for Computational Linguistics.

\bibitem[{Frermann and Klementiev(2019)}]{frermann2019inducing}
Lea Frermann and Alexandre Klementiev. 2019.
\newblock \href {https://doi.org/10.18653/v1/P19-1630} {Inducing {{Document
  Structure}} for {{Aspect}}-based {{Summarization}}}.
\newblock In \emph{Proceedings of the 57th {{Annual Meeting}} of the
  {{Association}} for {{Computational Linguistics}}}, pages 6263--6273,
  {Florence, Italy}. {Association for Computational Linguistics}.

\bibitem[{Gorinski and Lapata(2015)}]{gorinski2015movie}
Philip~John Gorinski and Mirella Lapata. 2015.
\newblock \href {https://doi.org/10.3115/v1/N15-1113} {Movie script
  summarization as graph-based scene extraction}.
\newblock In \emph{Proceedings of the 2015 Conference of the North {A}merican
  Chapter of the Association for Computational Linguistics: Human Language
  Technologies}, pages 1066--1076, Denver, Colorado. Association for
  Computational Linguistics.

\bibitem[{Grusky et~al.(2018)Grusky, Naaman, and Artzi}]{grusky2018newsroom}
Max Grusky, Mor Naaman, and Yoav Artzi. 2018.
\newblock \href {https://doi.org/10.18653/v1/N18-1065} {{N}ewsroom: A dataset
  of 1.3 million summaries with diverse extractive strategies}.
\newblock In \emph{Proceedings of the 2018 Conference of the North {A}merican
  Chapter of the Association for Computational Linguistics: Human Language
  Technologies, Volume 1 (Long Papers)}, pages 708--719, New Orleans,
  Louisiana. Association for Computational Linguistics.

\bibitem[{Kang et~al.(2018)Kang, Ammar, Dalvi, van Zuylen, Kohlmeier, Hovy, and
  Schwartz}]{kang2018dataset}
Dongyeop Kang, Waleed Ammar, Bhavana Dalvi, Madeleine van Zuylen, Sebastian
  Kohlmeier, Eduard Hovy, and Roy Schwartz. 2018.
\newblock \href {https://doi.org/10.18653/v1/N18-1149} {A dataset of peer
  reviews ({P}eer{R}ead): Collection, insights and {NLP} applications}.
\newblock In \emph{Proceedings of the 2018 Conference of the North {A}merican
  Chapter of the Association for Computational Linguistics: Human Language
  Technologies, Volume 1 (Long Papers)}, pages 1647--1661, New Orleans,
  Louisiana. Association for Computational Linguistics.

\bibitem[{Kedzie et~al.(2018)Kedzie, McKeown, and
  Daum{\'e}~III}]{kedzie2018content}
Chris Kedzie, Kathleen McKeown, and Hal Daum{\'e}~III. 2018.
\newblock \href {https://doi.org/10.18653/v1/D18-1208} {Content selection in
  deep learning models of summarization}.
\newblock In \emph{Proceedings of the 2018 Conference on Empirical Methods in
  Natural Language Processing}, pages 1818--1828, Brussels, Belgium.
  Association for Computational Linguistics.

\bibitem[{Krishna and Srinivasan(2018)}]{krishna2018generating}
Kundan Krishna and Balaji~Vasan Srinivasan. 2018.
\newblock \href {https://doi.org/10.18653/v1/N18-1153} {Generating
  {{Topic}}-{{Oriented Summaries Using Neural Attention}}}.
\newblock In \emph{Proceedings of the 2018 {{Conference}} of the {{North
  American Chapter}} of the {{Association}} for {{Computational Linguistics}}:
  {{Human Language Technologies}}, {{Volume}} 1 ({{Long Papers}})}, pages
  1697--1705, {New Orleans, Louisiana}. {Association for Computational
  Linguistics}.

\bibitem[{Lebanoff et~al.(2018)Lebanoff, Song, and Liu}]{lebanoff2018adapting}
Logan Lebanoff, Kaiqiang Song, and Fei Liu. 2018.
\newblock \href {https://doi.org/10.18653/v1/D18-1446} {Adapting the {{Neural
  Encoder}}-{{Decoder Framework}} from {{Single}} to {{Multi}}-{{Document
  Summarization}}}.
\newblock In \emph{Proceedings of the 2018 {{Conference}} on {{Empirical
  Methods}} in {{Natural Language Processing}}}, pages 4131--4141, {Brussels,
  Belgium}. {Association for Computational Linguistics}.

\bibitem[{Lin(2004)}]{lin2004rouge}
Chin-Yew Lin. 2004.
\newblock \href {https://www.aclweb.org/anthology/W04-1013} {{ROUGE}: A package
  for automatic evaluation of summaries}.
\newblock In \emph{Text Summarization Branches Out}, pages 74--81, Barcelona,
  Spain. Association for Computational Linguistics.

\bibitem[{Liu et~al.(2018)Liu, Saleh, Pot, Goodrich, Sepassi, Kaiser, and
  Shazeer}]{liu2018generating}
Peter~J. Liu, Mohammad Saleh, Etienne Pot, Ben Goodrich, Ryan Sepassi, Lukasz
  Kaiser, and Noam Shazeer. 2018.
\newblock \href {http://arxiv.org/abs/1801.10198} {Generating {{Wikipedia}} by
  {{Summarizing Long Sequences}}}.
\newblock \emph{arXiv:1801.10198 [cs]}.
\newblock ICLR.

\bibitem[{Liu and Lapata(2019{\natexlab{a}})}]{liu2019hierarchical}
Yang Liu and Mirella Lapata. 2019{\natexlab{a}}.
\newblock \href {https://doi.org/10.18653/v1/P19-1500} {Hierarchical
  transformers for multi-document summarization}.
\newblock In \emph{Proceedings of the 57th Annual Meeting of the Association
  for Computational Linguistics}, pages 5070--5081, Florence, Italy.
  Association for Computational Linguistics.

\bibitem[{Liu and Lapata(2019{\natexlab{b}})}]{liu2019text}
Yang Liu and Mirella Lapata. 2019{\natexlab{b}}.
\newblock \href {https://doi.org/10.18653/v1/D19-1387} {Text summarization with
  pretrained encoders}.
\newblock In \emph{Proceedings of the 2019 Conference on Empirical Methods in
  Natural Language Processing and the 9th International Joint Conference on
  Natural Language Processing (EMNLP-IJCNLP)}, pages 3730--3740, Hong Kong,
  China. Association for Computational Linguistics.

\bibitem[{Liu et~al.(2019)Liu, Ott, Goyal, Du, Joshi, Chen, Levy, Lewis,
  Zettlemoyer, and Stoyanov}]{liu2019roberta}
Yinhan Liu, Myle Ott, Naman Goyal, Jingfei Du, Mandar Joshi, Danqi Chen, Omer
  Levy, Mike Lewis, Luke Zettlemoyer, and Veselin Stoyanov. 2019.
\newblock \href {http://arxiv.org/abs/1907.11692} {{{RoBERTa}}: {{A Robustly
  Optimized BERT Pretraining Approach}}}.
\newblock \emph{arXiv:1907.11692 [cs]}.

\bibitem[{Lu et~al.(2009)Lu, Zhai, and Sundaresan}]{lu2009rated}
Yue Lu, ChengXiang Zhai, and Neel Sundaresan. 2009.
\newblock \href {https://doi.org/10.1145/1526709.1526728} {Rated aspect
  summarization of short comments}.
\newblock In \emph{Proceedings of the 18th International Conference on
  {{World}} Wide Web - {{WWW}} '09}, page 131, {Madrid, Spain}. {ACM Press}.

\bibitem[{Mihalcea and Tarau(2004)}]{mihalcea2004textrank}
Rada Mihalcea and Paul Tarau. 2004.
\newblock \href {https://www.aclweb.org/anthology/W04-3252} {{T}ext{R}ank:
  Bringing order into text}.
\newblock In \emph{Proceedings of the 2004 Conference on Empirical Methods in
  Natural Language Processing}, pages 404--411, Barcelona, Spain. Association
  for Computational Linguistics.

\bibitem[{Nallapati et~al.(2016)Nallapati, Zhou, dos Santos, Gul{\c{c}}ehre,
  and Xiang}]{nallapati2016abstractive}
Ramesh Nallapati, Bowen Zhou, Cicero dos Santos, {\c{C}}a{\u{g}}lar
  Gul{\c{c}}ehre, and Bing Xiang. 2016.
\newblock \href {https://doi.org/10.18653/v1/K16-1028} {Abstractive text
  summarization using sequence-to-sequence {RNN}s and beyond}.
\newblock In \emph{Proceedings of The 20th {SIGNLL} Conference on Computational
  Natural Language Learning}, pages 280--290, Berlin, Germany. Association for
  Computational Linguistics.

\bibitem[{Nenkova et~al.(2006)Nenkova, Vanderwende, and
  McKeown}]{nenkova2006compositional}
Ani Nenkova, Lucy Vanderwende, and Kathleen McKeown. 2006.
\newblock \href {https://dl.acm.org/doi/10.1145/1148170.1148269} {A
  compositional context sensitive multi-document summarizer: exploring the
  factors that influence summarization}.
\newblock In \emph{Proceedings of the 29th annual international ACM SIGIR
  conference on Research and development in information retrieval}, pages
  573--580. ACM.

\bibitem[{{Perez-Beltrachini} et~al.(2019){Perez-Beltrachini}, Liu, and
  Lapata}]{perez-beltrachini2019generating}
Laura {Perez-Beltrachini}, Yang Liu, and Mirella Lapata. 2019.
\newblock \href {https://www.aclweb.org/anthology/P19-1504} {Generating
  {{Summaries}} with {{Topic Templates}} and {{Structured Convolutional
  Decoders}}}.
\newblock In \emph{Proceedings of the 57th {{Annual Meeting}} of the
  {{Association}} for {{Computational Linguistics}}}, pages 5107--5116,
  {Florence, Italy}. {Association for Computational Linguistics}.

\bibitem[{Sauper and Barzilay(2009)}]{sauper2009automatically}
Christina Sauper and Regina Barzilay. 2009.
\newblock \href {https://www.aclweb.org/anthology/P09-1024} {Automatically
  {{Generating Wikipedia Articles}}: {{A Structure}}-{{Aware Approach}}}.
\newblock In \emph{Proceedings of the {{Joint Conference}} of the 47th {{Annual
  Meeting}} of the {{ACL}} and the 4th {{International Joint Conference}} on
  {{Natural Language Processing}} of the {{AFNLP}}}, pages 208--216, {Suntec,
  Singapore}. {Association for Computational Linguistics}.

\bibitem[{Titov and McDonald(2008)}]{titov2008joint}
Ivan Titov and Ryan McDonald. 2008.
\newblock \href {https://www.aclweb.org/anthology/P08-1036} {A {{Joint Model}}
  of {{Text}} and {{Aspect Ratings}} for {{Sentiment Summarization}}}.
\newblock In \emph{Proceedings of {{ACL}}-08: {{HLT}}}, pages 308--316,
  {Columbus, Ohio}. {Association for Computational Linguistics}.

\bibitem[{Wang and Ling(2016)}]{wang2016neural}
Lu~Wang and Wang Ling. 2016.
\newblock \href {https://doi.org/10.18653/v1/N16-1007} {Neural
  {{Network}}-{{Based Abstract Generation}} for {{Opinions}} and
  {{Arguments}}}.
\newblock In \emph{Proceedings of the 2016 {{Conference}} of the {{North
  American Chapter}} of the {{Association}} for {{Computational Linguistics}}:
  {{Human Language Technologies}}}, pages 47--57, {San Diego, California}.
  {Association for Computational Linguistics}.

\bibitem[{Wolf et~al.(2019)Wolf, Debut, Sanh, Chaumond, Delangue, Moi, Cistac,
  Rault, Louf, Funtowicz, and Brew}]{Wolf2019HuggingFacesTS}
Thomas Wolf, Lysandre Debut, Victor Sanh, Julien Chaumond, Clement Delangue,
  Anthony Moi, Pierric Cistac, Tim Rault, R'emi Louf, Morgan Funtowicz, and
  Jamie Brew. 2019.
\newblock \href {https://arxiv.org/abs/1910.03771} {Huggingface's transformers:
  State-of-the-art natural language processing}.
\newblock \emph{ArXiv}, abs/1910.03771.

\bibitem[{Yang et~al.(2018)Yang, Qu, Shen, Liu, Zhao, and Zhu}]{yang2018aspect}
Min Yang, Qiang Qu, Ying Shen, Qiao Liu, Wei Zhao, and Jia Zhu. 2018.
\newblock \href {https://www.aclweb.org/anthology/C18-1095} {Aspect and
  sentiment aware abstractive review summarization}.
\newblock In \emph{Proceedings of the 27th International Conference on
  Computational Linguistics}, pages 1110--1120, Santa Fe, New Mexico, USA.
  Association for Computational Linguistics.

\bibitem[{Yasunaga et~al.(2017)Yasunaga, Zhang, Meelu, Pareek, Srinivasan, and
  Radev}]{yasunaga2017graphbased}
Michihiro Yasunaga, Rui Zhang, Kshitijh Meelu, Ayush Pareek, Krishnan
  Srinivasan, and Dragomir Radev. 2017.
\newblock \href {https://doi.org/10.18653/v1/K17-1045} {Graph-based {{Neural
  Multi}}-{{Document Summarization}}}.
\newblock In \emph{Proceedings of the 21st {{Conference}} on {{Computational
  Natural Language Learning}} ({{CoNLL}} 2017)}, pages 452--462, {Vancouver,
  Canada}. {Association for Computational Linguistics}.

\end{thebibliography}
\bibliographystyle{acl_natbib}
\clearpage

\begin{appendices}
\section{Domain Statistics}
\label{app:domain_stats}

\begin{table}[H]
    \centering
    \begin{tabular}{@{}lrrr@{}}
    \toprule
    Domain & Train & Valid & Test \\\midrule
    Album & 24434 & 3104 & 3038 \\
    Animal & 16540 & 2005 & 2007 \\
    Artist & 26754 & 3194 & 3329 \\
    Building & 20449 & 2607 & 2482 \\
    Company & 24353 & 2946 & 3029 \\
    EducationalInstitution & 17634 & 2141 & 2267 \\
    Event & 6475 & 807 & 828 \\
    Film & 32129 & 4014 & 3981 \\
    Group & 11966 & 1462 & 1444 \\
    HistoricPlace & 4919 & 601 & 600 \\
    Infrastructure & 17226 & 1984 & 2091 \\
    MeanOfTransportation & 9277 & 1215 & 1170 \\
    OfficeHolder & 18177 & 2218 & 2333 \\
    Plant & 6107 & 786 & 774 \\
    Single & 14217 & 1734 & 1712 \\
    SoccerPlayer & 17599 & 2150 & 2280 \\
    Software & 13516 & 1637 & 1638 \\
    TelevisionShow & 8717 & 1128 & 1072 \\
    Town & 14818 & 1911 & 1831 \\
    WrittenWork & 15065 & 1843 & 1931 \\
    \bottomrule
    \end{tabular}
    \caption{\label{tab:datastats}The list of domains and the number of Wikipedia articles in each domain that contain at least one salient aspect.}
\end{table}

\section{Additional Samples}
\label{app:samples}

\begin{table}[htbp]
    \small
    \centering
    \begin{tabular}{@{}p{\linewidth}@{}}
        \toprule
        \textbf{Title: Recomposed by Max Richter: Vivaldi – The Four Seasons}\\\midrule
        Aspect: \textit{Critical Reception}\\\midrule
        \textbf{Gold}: recomposed by max richter  :  vivaldi  -  the four seasons received widespread acclaim from contemporary classical music critics  .  ivan hewett of the telegraph gave the album a very positive review  ,  stating  , "  as you would expect of a composer who once studied with the great modernist luciano berio  ,  richter is very self  -  aware  .$\ldots$\\
        \textbf{Ext.}: listen to recomposed by max richter : vivaldi , the four seasons now . i am highly impressed with ‘ recomposed ’ . the music then propels the audience into an atmosphere of isolation ; a delicate harmony that is sustained whilst hope takes centre stage . $\ldots$ \\
        \textbf{Abs.}: the allmusic review by michael g . nastos awarded the album 4 stars stating `` this is an album that generally considered for fans of the genre `` . $\ldots$\\
        \bottomrule
    \end{tabular}
    \caption{\label{tab:app_summaries1}Generated summaries from \textbf{Album} domain.}
\end{table}
\begin{table}[htb]
    \small
    \centering
    \begin{tabular}{@{}p{\linewidth}@{}}
        \toprule
        \textbf{Title: Pride and Glory (film)}\\\midrule
        Aspect: \textit{Plot}\\\midrule
        \textbf{Gold}: assistant chief francis tierney sr  .  is the head of a multigenerational new york city police department  (  nypd  )  family  ,  which includes his sons francis  "  franny  "  jr  . ,  ray  ,  and his son  -  in  -  law jimmy egan  .  deputy inspector franny is the commanding officer of the 31st precinct  ,  where sergeant jimmy is a patrol officer  ,  $\ldots$\\
        \textbf{Ext.}: as we know , under the macho code , this means that after two people who love each other end up beaten and bloody , they will somehow arrive at a catharsis . the plot involves how and why the four cops were killed . a family of police officers - patriarch , two sons , and a son - in - law - deals with corruption in a precinct in washington heights . $\ldots$\\
        \textbf{Abs.}: in the year before the events of the first film , the movie takes place in washington heights , d .  c .  , a .  army sergeant - in - law , ray ' s wife , and sister abby , living in washington city .  they have a romantic relationship with one of their officers .  while the four officers are called to `` the mental patient `` , $\ldots$\\
        \bottomrule
    \end{tabular}
    \caption{\label{tab:app_summaries2}Generated summaries from \textbf{Film} domain.}
\end{table}
\begin{table}[htb]
    \small
    \centering
    \begin{tabular}{@{}p{\linewidth}@{}}
        \toprule
        \textbf{Title: Dimitri Soudas}\\\midrule
        Aspect: \textit{Career}\\\midrule
        \textbf{Gold}: soudas served for one term as a school trustee at the western quebec school board from 2002 to 2005  .  between 2006 and 2011  ,  soudas was a  "  high profile  "  member of prime minister stephen harper  '  s communication team  ,  and one of the prime minister  '  s  "  closest and most faithful aides  . "  initially serving as a press secretary and later as an associate director of communications for the prime minister  '  s office  ,  $\ldots$ \\
        \textbf{Ext.}: april 2010 – after serving as a press secretary in the prime minister ’ s office , soudas was promoted to director of communications . " to fulfil the opportunities afforded by social media , directors of communication need to be aware of this trend and engage with it , " dimitri soudas writes in his master ' s thesis , a copy of which has been obtained by cbc news . $\ldots$\\
        \textbf{Abs.}: in 2001 , he was elected to the canadian house of commons as a member of the people ' s action party ( pc ) for the riding of yorkshire . he was re - elected in 2002 and 2006 . in 2006 , he was .\\
        \bottomrule
    \end{tabular}
    \caption{\label{tab:app_summaries3}Generated summaries from \textbf{OfficeHolder} domain.}
\end{table}

\clearpage
\section{Aspect Statistics}
\label{app:aspect_stats}
Table~\ref{tab:aspstats_all} and \ref{tab:aspstats_all2} shows aspect frequency statistics. Perf., hist., dist., ext., desc., dev., edu., nm., and intl. correspond to performance, history, distribution, extracurricular, description, development, education, naming, and international, respectively.

\begin{table}[htbp]
    \centering
    \small
    \begin{tabular}{@{}lr|lr@{}}
        \hline
        \multicolumn{2}{c}{ Album } & \multicolumn{2}{c}{ Animal } \\\hline
        reception & 11782 & description & 12729 \\
        critical reception & 6682 & distribution & 7813 \\
        background & 6202 & dist. \& habitat & 2967 \\
        commercial perf. & 2398 & taxonomy & 2737 \\
        release & 2209 & habitat & 2208 \\
        chart positions & 1891 & behavior & 2167 \\
        recording & 1490 & ecology & 1777 \\
        promotion & 1150 & diet & 1363 \\
        history & 1045 & reproduction & 1291 \\
        overview & 840 & biology & 1238 \\\hline
        \multicolumn{2}{c}{ Artist } & \multicolumn{2}{c}{ Building } \\\hline
        career & 10193 & history & 16885 \\
        biography & 8292 & architecture & 3223 \\
        early life & 7587 & desc. \& hist. & 1395 \\
        personal life & 6775 & description & 1382 \\
        music career & 2829 & location & 906 \\
        death & 1607 & interior & 877 \\
        life and career & 1512 & construction & 862 \\
        early life \& edu. & 1239 & exterior & 746 \\
        early years & 1129 & design & 623 \\
        exhibitions & 1030 & facilities & 572 \\\hline
        \multicolumn{2}{c}{ Company } & \multicolumn{2}{c}{ EducationalInstitution } \\\hline
        history & 21488 & history & 12798 \\
        products & 2921 & athletics & 5602 \\
        operations & 1630 & academics & 4638 \\
        services & 1019 & campus & 2471 \\
        controversy & 920 & sports & 1433 \\
        overview & 891 & student life & 1327 \\
        background & 572 & ext. activities & 1227 \\
        subsidiaries & 556 & curriculum & 1191 \\
        company history & 504 & facilities & 1189 \\
        technology & 471 & rankings & 836 \\\hline
        \multicolumn{2}{c}{ Event } & \multicolumn{2}{c}{ Film } \\\hline
        background & 3453 & plot & 25772 \\
        aftermath & 2483 & reception & 14003 \\
        history & 1361 & production & 13882 \\
        battle & 1228 & release & 7299 \\
        format & 461 & box office & 4572 \\
        prelude & 450 & critical reception & 4195 \\
        event & 416 & critical response & 2802 \\
        report & 323 & synopsis & 2626 \\
        summary & 321 & home media & 2461 \\
        casualties & 290 & filming & 2013 \\
    \hline
    \end{tabular}
    \vspace{-3mm}
    \caption{\label{tab:aspstats_all}Aspect frequency for 8 domains.}
\end{table}

\begin{table}[htb]
    \centering
    \small
    \begin{tabular}{@{}lr|lr@{}}
        \hline
        \multicolumn{2}{c}{ Group } & \multicolumn{2}{c}{ HistoricPlace } \\\hline
        history & 8894 & history & 3232 \\
        biography & 1206 & description & 1398 \\
        career & 1102 & desc. \& hist. & 1250 \\
        musical style & 683 & heritage listing & 942 \\
        background & 581 & architecture & 549 \\
        formation & 408 & location & 161 \\
        early years & 279 & historic uses & 90 \\
        legacy & 272 & preservation & 84 \\
        style & 265 & geography & 75 \\
        influences & 204 & interior & 70 \\\hline
        \multicolumn{2}{c}{ MeanOfTransportation } & \multicolumn{2}{c}{ OfficeHolder } \\\hline
        history & 2572 & personal life & 5119 \\
        design & 2152 & political career & 4950 \\
        operational hist. & 1989 & early life & 4740 \\
        design \& dev. & 1566 & career & 4115 \\
        service history & 1435 & biography & 2801 \\
        development & 1096 & education & 2168 \\
        construction & 933 & background & 1578 \\
        fate & 632 & death & 1402 \\
        background & 604 & legacy & 889 \\
        description & 602 & early life \& career & 859 \\\hline
        \multicolumn{2}{c}{ Plant } & \multicolumn{2}{c}{ Single } \\\hline
        description & 4684 & music video & 9606 \\
        dist. \& habitat & 1649 & critical reception & 3829 \\
        uses & 1585 & background & 3459 \\
        distribution & 1399 & reception & 2097 \\
        cultivation & 1387 & composition & 1729 \\
        taxonomy & 1121 & cover versions & 1594 \\
        ecology & 884 & content & 1266 \\
        conservation & 554 & release & 1045 \\
        etymology & 389 & commercial perf. & 849 \\
        taxonomy \& nm. & 384 & live performance & 113 \\\hline
        \multicolumn{2}{c}{ SoccerPlayer } & \multicolumn{2}{c}{ TelevisionShow } \\\hline
        intl. career & 8055 & plot & 2902 \\
        club career & 8029 & production & 2648 \\
        career & 6386 & reception & 2643 \\
        personal life & 3621 & synopsis & 1304 \\
        playing career & 1930 & premise & 944 \\
        early career & 1578 & history & 908 \\
        early life & 1191 & format & 842 \\
        professional & 992 & broadcast & 779 \\
        style of play & 887 & overview & 650 \\
        football career & 550 & critical reception & 583 \\\hline
        \multicolumn{2}{c}{ Town } & \multicolumn{2}{c}{ WrittenWork } \\\hline
        geography & 12667 & plot & 5495 \\
        demographics & 10949 & reception & 4970 \\
        history & 7298 & plot summary & 3900 \\
        education & 2868 & history & 2527 \\
        government & 1910 & background & 1218 \\
        2010 census & 1363 & adaptations & 1173 \\
        2000 census & 1284 & critical reception & 933 \\
        transportation & 1239 & manga & 830 \\
        economy & 1066 & history and profile & 803 \\
        name and history & 1002 & anime & 714 \\
        \hline
        \end{tabular}
        \vspace{-3mm}
        \caption{\label{tab:aspstats_all2}Aspect frequency for 10 domains.} %
\end{table}

\end{appendices}

\end{document}